\documentclass{article}
\usepackage{spconf,amsmath,amssymb,graphicx} 
\usepackage{multirow,booktabs}                
\usepackage{tabularx}                         
\usepackage{float}                            
\usepackage{xcolor}                           
\usepackage{hyperref}                         

\title{TimeGMM: Single-Pass Probabilistic Forecasting via Adaptive Gaussian Mixture Models with Reversible Normalization}
\name{Lei Liu\(^1\), Tengyuan Liu\(^1\), Hongwei Zhao\(^{1,\star}\), Jiahui Huang\(^1\), Ruibo Guo\(^1\), Bin Li\(^1\)\thanks{\(^\star\)Corresponding authors. The source code is available as: \url{https://github.com/USTC-AI4EEE/TimeGMM}}}
\address{\(^1\)University of Science and Technology of China, Hefei, 230026, China\\
\{liulei13, hwzhao, binli\}@ustc.edu.cn, \{tengyuanliu, jiahuihuang, ruiboguo\}@mail.ustc.edu.cn
}
\begin{document}
\ninept
\maketitle

\begin{abstract}
Probabilistic time series forecasting is crucial for quantifying future uncertainty, with significant applications in fields such as energy and finance. However, existing methods often rely on computationally expensive sampling or restrictive parametric assumptions to characterize future distributions, which limits predictive performance and introduces distributional mismatch. To address these challenges, this paper presents TimeGMM, a novel probabilistic forecasting framework based on Gaussian Mixture Models (GMM) that captures complex future distributions in a single forward pass. A key component is GMM-adapted Reversible Instance Normalization (GRIN), a novel module designed to dynamically adapt to temporal-probabilistic distribution shifts. The framework integrates a dedicated Temporal Encoder (TE-Module) with a Conditional Temporal-Probabilistic Decoder (CTPD-Module) to jointly capture temporal dependencies and mixture distribution parameters. Extensive experiments demonstrate that TimeGMM consistently outperforms state-of-the-art methods, achieving maximum improvements of 22.48\% in CRPS and 21.23\% in NMAE.
\end{abstract}
\section{Introduction}
\label{sec:introduction}

In recent years, time series analysis based on deep learning has witnessed rapid development, achieving remarkable progress in core tasks such as anomaly detection\cite{wu2024catch,miao2025parameter}, classification\cite{xu2023fits,campos2023lightts}, imputation\cite{yu2025ginar+,gao2025ssd}, and prediction\cite{liu2025timecma,chen2024pathformer,liu2024interpretable}. Against this backdrop, Probabilistic Time Series Forecasting (PTSF) has emerged as a key technology. By modeling the probability distribution of future outcomes, PTSF enables effective uncertainty quantification, risk mitigation, and informed planning, playing a crucial role in critical applications such as physical system\cite{liu2023segno}, economics\cite{sezer2020financial}, climate science\cite{mudelsee2019trend}, and traffic prediction\cite{wu2024autocts++}. However, the inherent randomness and complex temporal dependencies of time series data impose strict requirements on PTSF models, necessitating strong temporal representation learning capabilities and distribution modeling abilities, which makes long-term probabilistic time series prediction particularly challenging.

\begin{figure}[t]
    \centering
    \includegraphics[width=8.5cm]{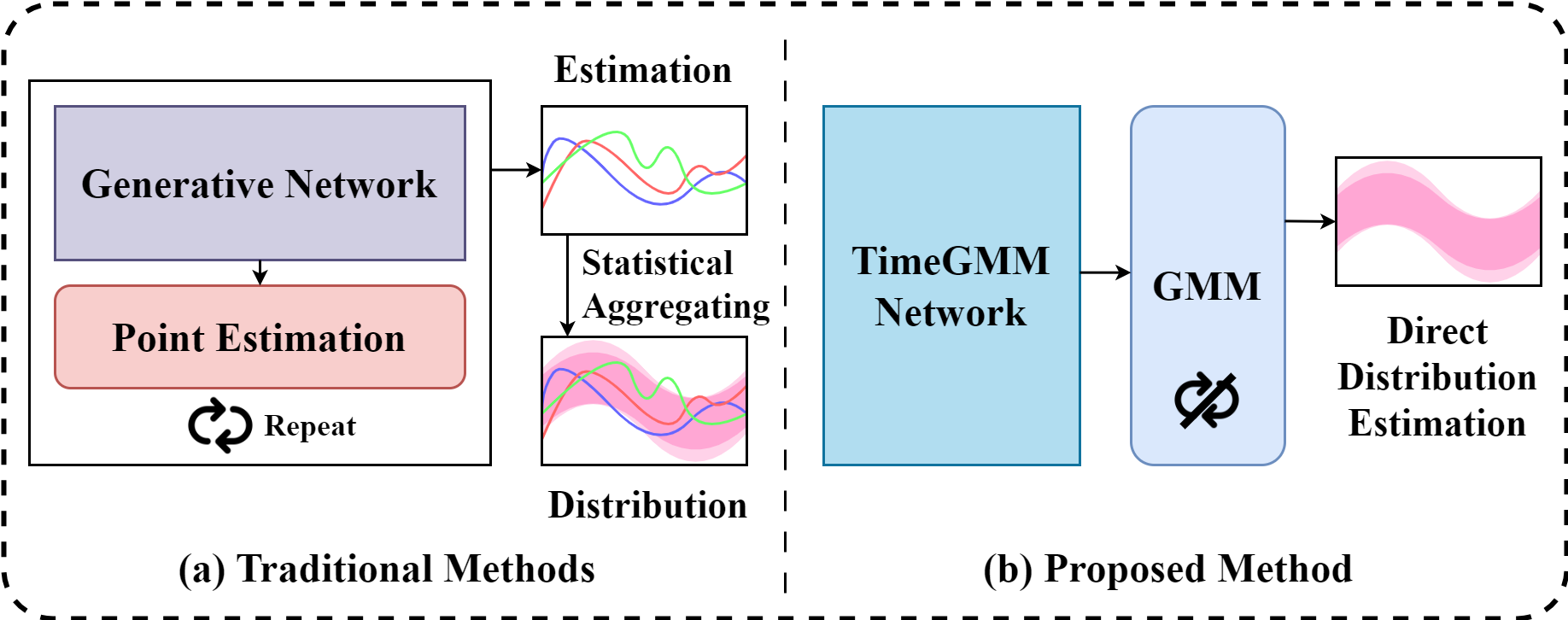}
    \vspace{-5pt}
    \caption{A comparison between the traditional methods and the proposed method.}
    \vspace{-10pt}
    \label{fig:prin}
\end{figure}

Most existing probabilistic time series forecasting methods rely on generative models, which can be broadly classified into three main categories: Variational Autoencoder (VAE)-based \cite{wu2025k,desai2021timevae}, diffusion-based\cite{kollovieh2023predict,li2022generative}, and flow-based\cite{rasul2020multivariate} approaches. As illustrated in Fig. \ref{fig:prin}, these methods infer probabilistic distributions by statistically aggregating multiple individual forecasts for each target time step. However, such strategies are essentially based on point estimation rather than direct probability modeling, and require multiple forward propagations to approximate the distribution, which leads to a decrease in the accuracy of probability distribution prediction due to the limited number of sampling steps.

In addition to the aforementioned methods, there are approaches such as DeepAR\cite{salinas2020deepar} that can directly estimate the probability distribution of time series. These models typically utilize predefined parameter distributions (such as Gaussian or Bernoulli distributions) to describe the target distribution. However, despite avoiding the repetitive sampling of point estimates, these methods still have significant limitations: the strong inductive bias introduced by the preset distributions often differs from the actual underlying data distribution, leading to performance degradation. Besides, such models are usually of relatively simple structure and lack specialized time modeling mechanisms, which limits their ability to learn complex temporal-probabilistic dependencies and ultimately leads to poor prediction accuracy.

To address these challenges, this paper proposes TimeGMM, a novel probabilistic forecasting method that eliminates the need for repeated sampling and avoids strong inductive biases associated with predefined parameter distributions. TimeGMM utilizes Gaussian Mixture Models to describe the target distribution, enabling flexible approximation of complex real-world data patterns and reducing performance degradation due to distribution mismatch. Additionally, this paper innovatively proposes a GMM-adapted Reversible Instance Normalization (GRIN) method that effectively addresses the issue of temporal-probabilistic distribution shift, and constructs an encoding-decoding architecture combining the Temporal Encoder (TE-Module) and Conditional Temporal-Probabilistic Decoder (CTPD-Module) to extract temporal dependencies and learn probability distribution parameters.

\begin{figure*}[ht]
    \centering
    \includegraphics[width=17cm]{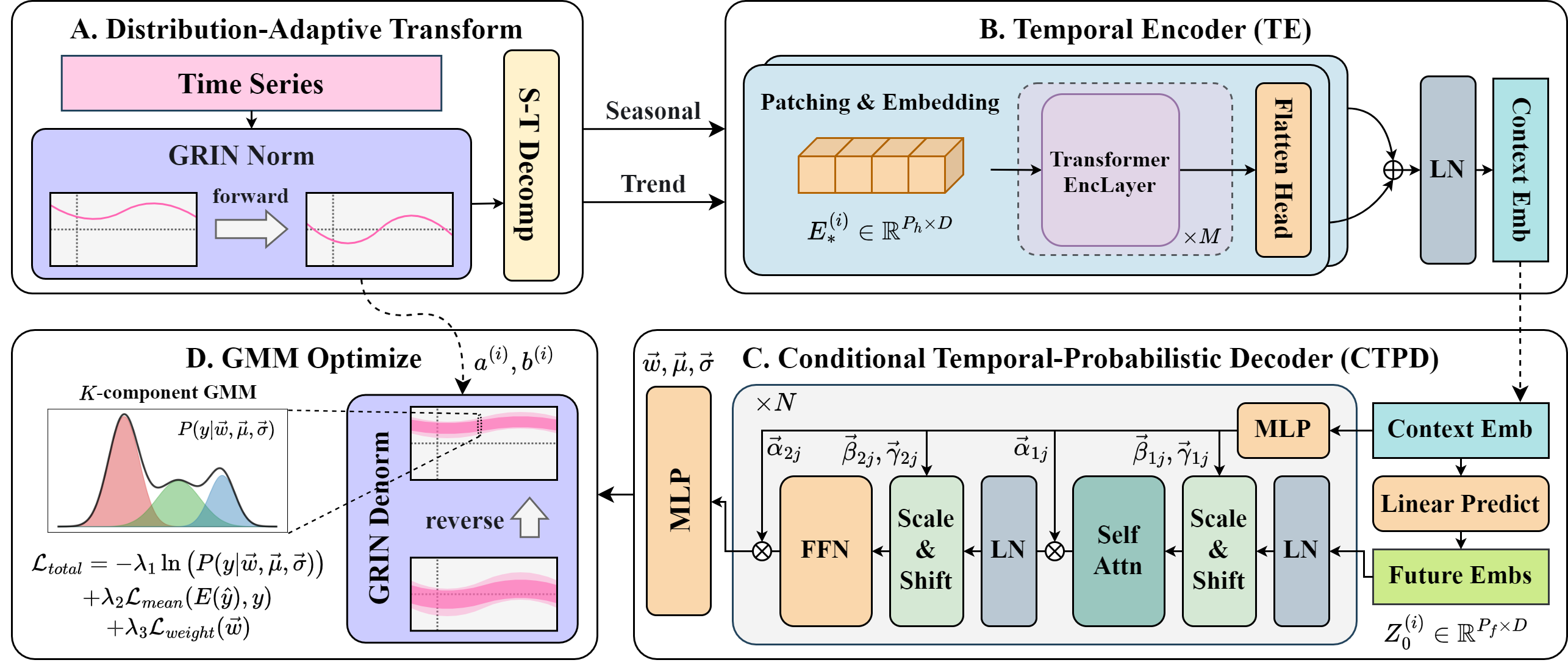}
    \vspace{-5pt}
    \caption{The overall framework of the proposed TimeGMM. \textbf{Part A} performs adaptive normalization and trend-seasonal decomposition; \textbf{Part B} captures temporal patterns, extracting and encoding historical information; \textbf{Part C} decodes encoder outputs into GMM parameters for probabilistic forecasting; \textbf{Part D} constructs the probability distribution from GMM parameters and computes the loss using negative log-likelihood, expected L2 error, and weight constraints.}
    \vspace{-10pt}
    \label{fig:arch}
\end{figure*}

The main contributions of this paper are summarized as follows:

\begin{itemize}
\item This paper proposes TimeGMM, a time series probabilistic prediction framework that generates flexible and asymmetric distribution estimates in a single forward pass, eliminating the need for repetitive sampling and mitigating distribution bias.

\item This paper proposes a GMM-adapted Reversible Instance Normalization (GRIN) module that adaptively normalizes the input while performing denormalization of the output based on the GMM parameterization mechanism, effectively mitigating the distribution shift problem in long-term probabilistic forecasting.

\item Experimental results demonstrate that the proposed method outperforms existing state-of-the-art approaches, achieving maximum improvements of 22.48\% and 21.23\% in CRPS and NMAE, respectively, over the second-best model.
\end{itemize}

\section{Methodology}
\label{sec:methodology}

The proposed TimeGMM in this paper consists of four components: data transformation, an encoder network, a decoder network, and a specially designed optimization objective. The overall framework is illustrated in Fig. \ref{fig:arch}.

\subsection{Distribution-Adaptive Transform}

To achieve more accurate and robust long-term probabilistic forecasting, inspired by RevIN\cite{kim2021reversible}, this paper proposes a reversible normalization method incorporating Gaussian Mixture Models, termed GRIN (GMM-adapted Reversible Instance Normalization), to address the distribution shift problem in probabilistic time series forecasting. Let \( x^{(i)}_t \) denote the value of the \(i\)-th variable at time \(t\), where \(i \in \{1, 2, \dots, V\}\) and \(t \in \{1, 2, \dots, L_h\}\). The entire multivariate time series with \(V\) variables and \(N\) time steps is represented as \(\mathbf{X} \in \mathbb{R}^{V \times L_h}\). The GRIN Norm transformation is defined as:

\begin{equation}\label{eqn1}
    \tilde{x}^{(i)}_t = a^{(i)} \left( \frac{x^{(i)}_t - \mathbb{E}_t\left[x^{(i)}\right]}{\sqrt{\mathrm{Var}_t\left[x^{(i)}\right] + \epsilon}} \right) + b^{(i)}
\end{equation}
where \(\tilde{x}^{(i)}_t\) denotes the normalized output, \(a^{(i)}, b^{(i)} \in \mathbb{R}\) are learnable parameters for each variable dimension, and \(\epsilon > 0\) is a small constant added for numerical stability.

Subsequently, seasonal-trend decomposition is introduced to improve the predictability of raw time series. The decomposition is performed using a sliding-window moving average approach, effectively separating trend and seasonal components. Formally, for a time series \(\tilde{X}^{(i)} \in \mathbb{R}^{L_h}\) of the \(i\)-th variable, the process is defined as:

\begin{equation}\label{eqn2}
        \tilde{X}^{(i)} _{T}, \tilde{X}^{(i)} _{S} = \mathrm{SeriesDecomp}\left(\tilde{X}^{(i)} \right)
\end{equation}
where \(\tilde{X}^{(i)}_{T}\) and \(\tilde{X}^{(i)}_{S}\) represent the trend and seasonal components, respectively.

\begin{table*}[t]
\vspace{-10pt}
\centering
\caption{Performance comparison of CRPS and NMAE on different datasets. The short dash (-) denotes runs that failed due to out-of-memory (OOM) errors.}
\label{tab1}
\vspace{4pt}
\renewcommand{\arraystretch}{0.9}
\begin{tabularx}{17cm}{X|c|cc|cc|cc|cc|cc}
\toprule[1.25pt]
\multirow{2}{*}{Dataset}     & Pred & \multicolumn{2}{c|}{TimeGMM} & \multicolumn{2}{c|}{\(K^2\)VAE} & \multicolumn{2}{c|}{CSDI} & \multicolumn{2}{c|}{GRU NVP} & \multicolumn{2}{c}{TimeGrad} \\
                             & len  & CRPS          & NMAE         & CRPS        & NMAE       & CRPS          & NMAE         & CRPS         & NMAE         & CRPS          & NMAE         \\ \midrule[1.25pt]
\multirow{4}{*}{ETTm1}       & 96   & \textcolor{red}{\textbf{0.206}} & \textcolor{red}{\textbf{0.267}} & \textcolor{blue}{\underline{0.232}} & \textcolor{blue}{\underline{0.284}} & 0.236 & 0.308 & 0.383 & 0.488 & 0.522 & 0.645 \\
                             & 192  & \textcolor{red}{\textbf{0.227}} & \textcolor{red}{\textbf{0.296}} & \textcolor{blue}{\underline{0.259}} & \textcolor{blue}{\underline{0.323}} & 0.291 & 0.377 & 0.396 & 0.514 & 0.603 & 0.748 \\
                             & 336  & \textcolor{red}{\textbf{0.245}} & \textcolor{red}{\textbf{0.321}} & \textcolor{blue}{\underline{0.262}} & \textcolor{blue}{\underline{0.330}} & 0.322 & 0.419 & 0.486 & 0.630 & 0.601 & 0.759 \\
                             & 720  & \textcolor{red}{\textbf{0.273}} & \textcolor{red}{\textbf{0.359}} & \textcolor{blue}{\underline{0.294}} & \textcolor{blue}{\underline{0.373}} & 0.448 & 0.578 & 0.546 & 0.707 & 0.621 & 0.793 \\ \midrule
\multirow{4}{*}{ETTm2}       & 96   & \textcolor{red}{\textbf{0.110}} & \textcolor{red}{\textbf{0.134}} & 0.126 & \textcolor{blue}{\underline{0.144}} & \textcolor{blue}{\underline{0.115}} & 0.146 & 0.319 & 0.413 & 0.427 & 0.525 \\
                             & 192  & \textcolor{red}{\textbf{0.131}} & \textcolor{red}{\textbf{0.159}} & 0.148 & \textcolor{blue}{\underline{0.170}} & \textcolor{blue}{\underline{0.147}} & 0.189 & 0.326 & 0.427 & 0.424 & 0.530 \\
                             & 336  & \textcolor{red}{\textbf{0.149}} & \textcolor{red}{\textbf{0.181}} & \textcolor{blue}{\underline{0.164}} & \textcolor{blue}{\underline{0.187}} & 0.190 & 0.248 & 0.449 & 0.580 & 0.469 & 0.566 \\
                             & 720  & \textcolor{red}{\textbf{0.173}} & \textcolor{red}{\textbf{0.208}} & \textcolor{blue}{\underline{0.221}} & \textcolor{blue}{\underline{0.275}} & 0.239 & 0.306 & 0.561 & 0.749 & 0.470 & 0.561 \\ \midrule
\multirow{4}{*}{ETTh1}       & 96   & \textcolor{red}{\textbf{0.258}} & \textcolor{red}{\textbf{0.327}} & \textcolor{blue}{\underline{0.264}} & \textcolor{blue}{\underline{0.336}} & 0.437 & 0.557 & 0.379 & 0.481 & 0.455 & 0.585 \\
                             & 192  & \textcolor{red}{\textbf{0.279}} & \textcolor{red}{\textbf{0.351}} & \textcolor{blue}{\underline{0.290}} & \textcolor{blue}{\underline{0.372}} & 0.496 & 0.625 & 0.425 & 0.531 & 0.516 & 0.680 \\
                             & 336  & \textcolor{red}{\textbf{0.292}} & \textcolor{red}{\textbf{0.365}} & \textcolor{blue}{\underline{0.308}} & \textcolor{blue}{\underline{0.394}} & 0.454 & 0.574 & 0.458 & 0.580 & 0.512 & 0.666 \\
                             & 720  & \textcolor{red}{\textbf{0.311}} & \textcolor{blue}{\underline{0.397}} & \textcolor{blue}{\underline{0.314}} & \textcolor{red}{\textbf{0.396}} & 0.528 & 0.657 & 0.502 & 0.643 & 0.523 & 0.672 \\ \midrule
\multirow{4}{*}{ETTh2}       & 96   & \textcolor{red}{\textbf{0.143}} & \textcolor{red}{\textbf{0.176}} & \textcolor{blue}{\underline{0.162}} & \textcolor{blue}{\underline{0.189}} & 0.164 & 0.214 & 0.432 & 0.548 & 0.358 & 0.448 \\
                             & 192  & \textcolor{red}{\textbf{0.161}} & \textcolor{red}{\textbf{0.198}} & \textcolor{blue}{\underline{0.186}} & \textcolor{blue}{\underline{0.213}} & 0.226 & 0.294 & 0.625 & 0.766 & 0.457 & 0.575 \\
                             & 336  & \textcolor{red}{\textbf{0.181}} & \textcolor{red}{\textbf{0.220}} & \textcolor{blue}{\underline{0.257}} & \textcolor{blue}{\underline{0.263}} & 0.274 & 0.353 & 0.793 & 0.942 & 0.481 & 0.606 \\
                             & 720  & \textcolor{red}{\textbf{0.186}} & \textcolor{red}{\textbf{0.227}} & \textcolor{blue}{\underline{0.280}} & \textcolor{blue}{\underline{0.278}} & 0.302 & 0.382 & 0.539 & 0.688 & 0.445 & 0.550 \\ \midrule
\multirow{4}{*}{Electricity} & 96   & \textcolor{red}{\textbf{0.067}} & \textcolor{red}{\textbf{0.084}} & \textcolor{blue}{\underline{0.073}} & \textcolor{blue}{\underline{0.093}} & 0.153 & 0.203 & 0.094 & 0.118 & 0.096 & 0.119 \\
                             & 192  & \textcolor{red}{\textbf{0.073}} & \textcolor{red}{\textbf{0.091}} & \textcolor{blue}{\underline{0.080}} & \textcolor{blue}{\underline{0.102}} & 0.200 & 0.264 & 0.097 & 0.121 & 0.100 & 0.124 \\
                             & 336  & \textcolor{blue}{\underline{0.080}} & \textcolor{red}{\textbf{0.100}} & \textcolor{red}{\textbf{0.054}} & \textcolor{blue}{\underline{0.107}} & -     & -     & 0.099 & 0.123 & 0.102 & 0.126 \\
                             & 720  & \textcolor{blue}{\underline{0.093}} & \textcolor{red}{\textbf{0.116}} & \textcolor{red}{\textbf{0.057}} & \textcolor{blue}{\underline{0.117}} & -     & -     & 0.114 & 0.144 & 0.108 & 0.134 \\ \midrule
\multirow{4}{*}{Weather}     & 96   & \textcolor{red}{\textbf{0.053}} & \textcolor{red}{\textbf{0.064}} & 0.080 & \textcolor{blue}{\underline{0.086}}  & \textcolor{blue}{\underline{0.068}} & 0.087 & 0.116 & 0.145 & 0.130 & 0.164 \\
                             & 192  & \textcolor{red}{\textbf{0.059}} & \textcolor{red}{\textbf{0.071}} & 0.079 & \textcolor{blue}{\underline{0.083}} & \textcolor{blue}{\underline{0.068}} & 0.086 & 0.122 & 0.147 & 0.127 & 0.158 \\
                             & 336  & \textcolor{red}{\textbf{0.059}} & \textcolor{red}{\textbf{0.072}} & \textcolor{blue}{\underline{0.082}} & \textcolor{blue}{\underline{0.093}} & 0.083 & 0.098 & 0.128 & 0.160 & 0.130 & 0.162 \\
                             & 720  & \textcolor{red}{\textbf{0.064}} & \textcolor{red}{\textbf{0.078}} & \textcolor{blue}{\underline{0.084}} & \textcolor{blue}{\underline{0.099}} & 0.087 & 0.102 & 0.110 & 0.135 & 0.113 & 0.136 \\ \midrule
\multirow{4}{*}{Exchange}    & 96   & \textcolor{red}{\textbf{0.020}} & \textcolor{red}{\textbf{0.024}} & 0.031 & \textcolor{blue}{\underline{0.032}} & \textcolor{blue}{\underline{0.028}} & 0.036 & 0.071 & 0.091 & 0.068 & 0.079 \\
                             & 192  & \textcolor{red}{\textbf{0.029}} & \textcolor{red}{\textbf{0.036}} & \textcolor{blue}{\underline{0.032}} & \textcolor{blue}{\underline{0.040}} & 0.045 & 0.058 & 0.068 & 0.087 & 0.087 & 0.100 \\
                             & 336  & \textcolor{red}{\textbf{0.043}} & \textcolor{red}{\textbf{0.052}} & \textcolor{blue}{\underline{0.048}} & \textcolor{blue}{\underline{0.054}} & 0.060 & 0.076 & 0.072 & 0.091 & 0.074 & 0.086 \\
                             & 720  & \textcolor{red}{\textbf{0.068}} & \textcolor{red}{\textbf{0.083}} & \textcolor{blue}{\underline{0.069}} & \textcolor{blue}{\underline{0.084}} & 0.143 & 0.173 & 0.079 & 0.103 & 0.099 & 0.113 \\ \midrule[1.25pt]
\multicolumn{2}{c|}{1st Count} & 26 & 27 & 2 & 1 & 0 & 0 & 0 & 0 & 0 & 0 \\
\bottomrule[1.25pt]
\end{tabularx}
\vspace{-10pt}
\end{table*}

\subsection{Temporal Encoder (TE)}

To better capture the temporal features, this paper constructs two branches with independent weights for the trend and seasonal components respectively in the encoder.

Specifically, to perceive the temporal input, TimeGMM first segments the time series into patches following \cite{nie2022time}, and then encodes each patch using an MLP layer to obtain the input embedding \(E^{(i)}_{*}\in\mathbb{R}^{P_h\times D}\), where \(*\) denotes the trend component (\(*=T\)) or the seasonal component (\(*=S\)), \(P_h\) denotes the number of patches from the historical input, and \(D\) denotes the embedding dimension.

\begin{equation}\label{eqn3}
    E^{(i)}_*=\mathrm{MLP}_{enc}\left(\mathrm{Patching}(\tilde{X}^{(i)} _*) \right)
\end{equation}

Then, to facilitate effective feature extraction, a Transformer encoder is employed to process the temporal encodings. Subsequently, a flattening layer is incorporated to project the Transformer's output into the embedding \(E'^{(i)}\), which can be formulated as follows:

\begin{equation}\label{eqn4}
E'^{(i)}_* = \mathrm{FlattenHead}\left(\mathrm{TransformerEncLayer}^M(E^{(i)}_*)\right)
\end{equation}
where $M$ indicates the total number of stacked encoder layers.

Finally, the embeddings output by the trend and seasonal branches are summed together and normalized via LayerNorm to obtain the representation encoding of the temporal input \(E^{(i)}_C\).

\subsection{Conditional Temporal-Probabilistic Decoder (CTPD)}

In this paper, the decoder is primarily designed to estimate the three parameters of the Gaussian Mixture Model: the weights $\vec{w}$, means $\vec{\mu}$, and standard deviations $\mathbf{\sigma}$, where $\vec{w}, \vec{\mu}, \vec{\sigma} \in \mathbb{R}^{V \times L_f \times K}$. Here, $L_f$ denotes the forecast horizon, and $K$ represents the number of Gaussian components in the mixture.

First, based on the representation \(E^{(i)}_C\) from the encoder, this paper applies a linear transformation \(\mathbf{W}_{pred}\in\mathbb{R}^{(P_f\times D)\times D}\) to obtain an initial approximation of the future prediction:

\begin{equation}\label{eqn6}
    Z^{(i)}_0=\mathbf{W}_{pred}E^{(i)}_C
\end{equation}

Second, inspired by recent advances in image generation \cite{peebles2023scalable}, this work incorporates the adaLN mechanism to conditionally modulate the Transformer's representations for refined prediction. The overall pipeline operates as follows:

\begin{equation}\label{eqn7}
\vec{\alpha}^{(i)}_{1j},\vec{\beta}^{(i)}_{1j},\vec{\gamma}^{(i)}_{1j},\vec{\alpha}^{(i)}_{2j},\vec{\beta}^{(i)}_{2j},\vec{\gamma}^{(i)}_{2j}=\mathrm{MLP}_j\left(E^{(i)}_C\right)
\end{equation}
\begin{equation}\label{eqn8}
Z'^{(i)}_j=\vec{\alpha}^{(i)}_{1j} \mathrm{SelfAttn}_j\left(\vec{\gamma}^{(i)}_{1j} \mathrm{LN}^{\#}(Z^{(i)}_{j-1}) +\vec{\beta}^{(i)}_{1j} \right)
\end{equation}

\begin{equation}\label{eqn9}
Z^{(i)}_j=\vec{\alpha}^{(i)}_{2j} \mathrm{FFN}_j\left(\vec{\gamma}^{(i)}_{2j} \mathrm{LN}^{\#}(Z'^{(i)}_j) +\vec{\beta}^{(i)}_{2j} \right)
\end{equation}
where $j\in\{1,2,\dots,N\}$ denotes the decoder layer index, $\vec{\alpha}^{(i)}_*$, $\vec{\beta}^{(i)}_*$, $\vec{\gamma}^{(i)}_* \in \mathbb{R}^{D}$ are learnable coefficients of i-th variable, and $\mathrm{LN}^{\#}$ denotes the LayerNorm layer with its learnable parameters removed.

Finally, a MLP layer is used to decode \(Z^{(i)}_N\) into three sets of GMM parameters: \(\vec{w}^{(i)}\), \(\vec{\mu}^{(i)}\), and \(\vec{\sigma}^{(i)}\):
\begin{equation}\label{eqn10}      \vec{w}^{(i)},\tilde{\vec{\mu}}^{(i)},\tilde{\vec{\sigma}}^{(i)}=\mathrm{MLP}_{dec}\left(Z^{(i)}_N\right)
\end{equation}
where \(\tilde{\vec{\mu}}^{(i)},\tilde{\vec{\sigma}}^{(i)}\) represent the mean and variance of the GMM distribution before GRIN denormalization.

\subsection{GMM Optimize}

To reconstruct the true probability distribution of the time series data, the first step is to de-normalize the normalized GMM parameters through GRIN Denorm:

\begin{equation}\label{eqn11}
\vec{\mu}^{(i)}=\sqrt{\mathrm{Var}_t\left[x^{(i)}\right]}\frac{\tilde{\vec{\mu}}^{(i)} -  b^{(i)}}{a^{(i)}+\epsilon}+\mathbb{E}_t\left[x^{(i)}\right]
\end{equation}
\begin{equation}\label{eqn12}
\vec{\sigma}^{(i)}=\sqrt{\mathrm{Var}_t\left[x^{(i)}\right]}\frac{\tilde{\vec{\sigma}}^{(i)}}{a^{(i)}+\epsilon}
\end{equation}

Then, a continuous probability distribution function can be constructed through the GMM parameters:

\begin{equation}\label{eqn14}
    P(y^{(i)}|\vec{w}^{(i)},\vec{\mu}^{(i)},\vec{\sigma}^{(i)})=\sum_{k=1}^{K}w^{(i)}_k \cdot N(y^{(i)}|\mu^{(i)}_k,\sigma^{(i)}_k)
\end{equation}

To optimize the GMM probability distribution, this paper introduces the negative log-likelihood loss as the main objective function based on the maximum likelihood estimation method, and its expression is as follows:

\begin{equation}\label{eqn15}
\mathcal{L}_{NLL}=-\ln\left(P(y|\vec{w},\vec{\mu}, \vec{\sigma})\right)
\end{equation}

To achieve improved performance, this paper additionally incorporates the expected loss \(\mathcal{L}_{mean}\) and the weight sum loss \(\mathcal{L}_{weight}\) into the overall loss function. The expected loss is designed to optimize the expected value of the probability distribution function, while the weight sum loss enforces soft constraints on the GMM mixture weights. Both components are computed using the L2 loss. By introducing the coefficient \(\lambda_1\), \(\lambda_2\), and \(\lambda_3\), the total loss function is constructed as follows:

\begin{equation}\label{eqn16}
\mathcal{L}_{total}=\lambda_1\mathcal{L}_{NLL}+\lambda_2\mathcal{L}_{mean}+\lambda_3\mathcal{L}_{weight}
\end{equation}

It should be noted that only during the testing phase are the weights $\vec{w}$ passed through a Softmax layer to enforce a strict summation-to-one constraint.

\section{Experiments}
\subsection{Experiment Settings}

To better evaluate the proposed TimeGMM, this paper follows the experimental setup provided by the ProbTS benchmark \cite{zhang2024probts} for testing and comparison. The detailed experimental configurations are described as follows:

\begin{table}[!ht]
\vspace{-10pt}
\centering
\caption{Basic information of the datasets.}
\label{tab2}
\vspace{4pt}
\renewcommand{\arraystretch}{0.8}
\begin{tabularx}{8cm}{X|cccc}
\toprule[1.5pt]
Dataset & \#var. & range. & freq. & timesteps \\
\midrule[1.5pt]
ETTm1/m2 & 7 & \(\mathbb{R}^+\) & 15min & 69,680 \\
ETTh1/h2 & 7 & \(\mathbb{R}^+\) & H & 17,420 \\
Electricity & 321 & \(\mathbb{R}^+\) & H & 26,304 \\
Weather & 21 & \(\mathbb{R}^+\) & 10min & 52,696 \\
Exchange & 8 & \(\mathbb{R}^+\) & Busi. Day & 7,588 \\
\bottomrule[1.5pt]
\end{tabularx}
\vspace{-10pt}
\end{table}

\textbf{Datasets}\quad This paper conducts experiments on 7 commonly used time series forecasting datasets: ETTm1/m2, ETTh1/h2, Electricity, Weather, and Exchange. The input length is set to 96, and the output lengths are set to \{96, 192, 336, 720\}. Detailed information about the datasets can be found in Table \ref{tab2}.

\textbf{Baselines}\quad To facilitate comprehensive performance evaluation, this study compares against several state-of-the-art probabilistic forecasting methods, including \textbf{VAE-based} method \(K^2\)VAE\cite{wu2025k}, \textbf{diffusion-based} methods CSDI\cite{tashiro2021csdi} and TimeGrad\cite{rasul2021autoregressive}, as well as \textbf{flow-based} approach GRU NVP\cite{rasul2020multivariate}. The performance data of the baselines are obtained from their official reports on the ProbTS benchmark\cite{zhang2024probts}.

\textbf{Implementation Details}\quad TimeGMM is optimized using the AdamW optimizer with a learning rate of $2.5 \times 10^{-4}$ and an L2 weight decay of 0.01. The presented values are the mean deviation of 3 independent experiments. All code is implemented with PyTorch, and experiments are conducted on a server with dual NVIDIA GeForce RTX 4090 GPUs.

\subsection{Main Results}

The Continuous Ranked Probability Score (CRPS) and Normalized Mean Absolute Error (NMAE) performance of all models on the benchmark are shown in Table \ref{tab1}, where \textcolor{red}{\textbf{red}} indicates the best performance and \textcolor{blue}{\underline{blue}} represents the second best.

As can be seen from the performance comparison, TimeGMM achieves the best results across all datasets (when considering all prediction horizons collectively), delivering accurate and robust long-term probabilistic forecasts and outperforming existing state-of-the-art models.

\begin{table}[!t]
\vspace{-10pt}
\centering
\caption{Ablation experimental results on four datasets.}
\label{tab3}
\vspace{4pt}
\renewcommand{\arraystretch}{0.9}
\begin{tabularx}{7.5cm}{X|X|>{\centering\arraybackslash}p{1.8cm}}
\toprule[1.5pt]
Dataset & Method & CRPS \\
\midrule[1.5pt]
\multirow{3}{*}{ETTm1} & TimeGMM & \textbf{0.2378} \\
& w/o GMM & 0.2407\\
& w/o GRIN & 0.2743\\
\midrule
\multirow{3}{*}{ETTm2} & TimeGMM & \textbf{0.1409} \\
& w/o GMM & 0.1440\\
& w/o GRIN & 0.1588\\
\midrule
\multirow{3}{*}{ETTh2} & TimeGMM & \textbf{0.1677} \\
& w/o GMM & 0.1681\\
& w/o GRIN & 0.2036\\
\midrule
\multirow{3}{*}{Weather} & TimeGMM & \textbf{0.0585} \\
& w/o GMM & 0.0613\\
& w/o GRIN & 0.0611\\
\bottomrule[1.5pt]
\end{tabularx}
\vspace{-10pt}
\end{table}
\subsection{Ablation Study}

Ablation studies on four datasets assess key components of TimeGMM: removing GMM (w/o GMM), and removing GRIN (w/o GRIN). With input length 96 and outputs {96, 192, 336, 720}, results are averaged over forecasting horizons and evaluated using CRPS and NMAE. Details are in Table \ref{tab3}.

It can be seen from the ablation results that the ablation of both GRIN and GMM has a significant impact on the model's probability prediction accuracy, indicating that both play a positive role in accurate and robust long-term probability prediction, which verifies the design of this paper.

\section{Conclusion}

This paper proposes a novel probabilistic time series forecasting framework, TimeGMM, based on Gaussian Mixture Model. By integrating GMM, this framework flexibly models complex distributions, avoids the inductive bias and mismatch from preset distributions, and performs probabilistic prediction in a single forward pass. To enhance the prediction robustness of the model, an GMM-adapted instance normalization module, GRIN, is designed to effectively alleviate the distribution shift phenomenon in probabilistic prediction. Through the encoder-decoder architecture composed of Temporal Encoder (TE-Module) and Conditional Temporal-Probabilistic Decoder (CTPD-Module), TimeGMM can collaboratively model the temporal dependency and probabilistic distribution parameters generation. Experimental results show that TimeGMM significantly outperforms existing state-of-the-art methods on multiple public benchmark datasets, achieving up to 22.48\% and 21.23\% performance improvements in CRPS and NMAE evaluation metrics. This study provides an effective and universal solution for high-precision and high-robustness time series probabilistic prediction, with promising practical application prospects.

\newpage
\bibliographystyle{IEEEbib}
\bibliography{refs}

\end{document}